\documentclass[12pt]{article}
\newcommand{\keywords}[1]{\textbf{Keywords:} #1}

\usepackage[utf8]{inputenc}    
\usepackage[T1]{fontenc}       
\usepackage{lmodern}           
\usepackage{bm}        
\usepackage{amsmath}   
\usepackage{accents}   
\usepackage[utf8]{inputenc}   
\usepackage[T1]{fontenc}  
\usepackage{algorithm}
\usepackage{float}
\usepackage{algpseudocode}
\usepackage{amsthm}     
\newtheorem{theorem}{Theorem}
\usepackage{soul}

\usepackage{varwidth}

\usepackage[a4paper, margin=1in]{geometry} 

\usepackage{amsmath, amssymb, amsfonts, amsthm}

\usepackage{graphicx}          
\usepackage{float}             
\usepackage{caption}           
\usepackage{subcaption}        

\usepackage{booktabs}          
\usepackage{array}             
\usepackage{tabularx}          
\usepackage{multirow}          
\usepackage{microtype}
\usepackage[colorlinks=true, linkcolor=blue, urlcolor=blue, citecolor=blue]{hyperref}
\usepackage{xcolor}            
\usepackage{placeins}

\usepackage{cleveref}          
\crefname{section}{Section}{Sections}
\crefname{figure}{Figure}{Figures}
\crefname{table}{Table}{Tables}
\crefname{equation}{Equation}{Equations}

\usepackage[bottom]{footmisc}  
\usepackage{enumitem}          
\usepackage{setspace} 
\usepackage{authblk}

\title{\normalsize \textbf{Discovering Governing Equations in the Presence of Uncertainty}}

\author[1]{\normalsize Ridwan Olabiyi}
\author[2]{Han Hu}
\author[1]{Ashif Iquebal}

\affil[1]{\normalsize School of Computing and Augmented Intelligence, Arizona State University}
\affil[2]{Department of Mechanical Engineering, University of Arkansas}
\date{}

\begin{document}

\maketitle

\small
\noindent\textbf{Abstract.} In the study of complex dynamical systems, understanding and accurately modeling the underlying physical processes is crucial for predicting system behavior and designing effective interventions. Yet real‑world systems exhibit pronounced input (or system) variability and are observed through noisy, limited data conditions that confound traditional discovery methods that assume fixed‑coefficient deterministic models. In this work, we theorize that accounting for system variability together with measurement noise is the key to consistently discover the governing equations underlying dynamical systems. As such, we introduce a stochastic inverse physics‑discovery (SIP) framework that treats the unknown coefficients as random variables and infers their posterior distribution by minimizing the Kullback–Leibler divergence between the push‑forward of the posterior samples and the empirical data distribution. Benchmarks on four canonical problems—the Lotka–Volterra predator–prey system (multi‑ and single‑trajectory), the historical Hudson Bay lynx–hare data, the chaotic Lorenz attractor, and fluid infiltration in porous media using low‑ and high‑viscosity liquids—show that SIP consistently identifies the correct equations and lowers coefficient root‑mean‑square error by an average of 82 \% relative to the Sparse Identification of Nonlinear Dynamics (SINDy) approach and its Bayesian variant. The resulting posterior distributions yield 95 \% credible intervals that closely track the observed trajectories, providing interpretable models with quantified uncertainty. SIP thus provides a robust, data‑efficient approach for consistent physics discovery in noisy, variable, and data‑limited~settings.


\small

\section{Introduction}
\label{sec:intro} 
Physical laws have been the cornerstone of understanding the behavior of complex systems and enabling scientific discoveries dating as far back as Archimedes and Ptolemy \cite{peters1915ptolemy, ptolemy2014almagest}. Back then, scientific discoveries were based on decades of systematic observation, empirical evidence, and meticulous experimentation \cite{westfall1983never}, and yet, many were still refuted \cite{kuhn1962structure}. In recent times, the ability to rapidly gather large datasets with inexpensive sensors and the availability of powerful computing resources is transforming how we discover the underlying dynamics of complex systems using statistical~and~machine~learning techniques.

The earliest works on the identification of dynamical systems from measured data came from Norbert Wiener and Andrey Kolmogorov on prediction theory and stochastic processes in the 1940s and 1950s \cite{kolmogorov1941stationary, wiener1949extrapolation}. A major breakthrough was reported in the works of Bongard, Schmidt, and Lispson \cite{bongard2007automated, schmidt2009distilling}, who formulated the physics identification problem as symbolic regression involving linear, non-linear and higher-order differential terms \cite{voss2004nonlinear, bongard2007automated, schmidt2009distilling}. 
However, poor scalability to large systems and predisposition to overfitting severely limited their broader applicability \cite{brunton2016discovering, brunton2020machine}. 
Building on these initial efforts, Brunton et al. \cite{Brunton2016} hypothesized that \textit{physical laws are often sparse over the space of linear and nonlinear functions, re-envisioning the dynamical systems discovery as a sparse system identification problem}. This hypothesis laid the foundation for the development of the Sparse Identification of Nonlinear Dynamics (SINDy) framework, which addressed some of the scalability issues in the previous approaches by employing sparse regression methods. Initial efforts focused on regularized regression methods such as LASSO and sequential threshold least squares to parsimoniously recover the unknown coefficients of the predefined physics terms \cite{brunton2016discovering}. While SINDy offers an efficient approach to parsimoniously identify the unknown dynamical system from data, thus enabling the extraction of interpretable models across different scenarios \cite{champion2020unified, champion2019data}, some critical challenges remained: (a) recovered equations either missed certain physics terms or erroneously learned additional higher-order terms, (b) the consistency of the estimated coefficients of the identified physics terms is not guaranteed, and (c) it generalizes poorly beyond the observed space of state variables (e.g., see Section \ref{sec:hudson_bay_dataset}, lynx-hare case study).

More recent research focused on Hidden Physics Models \cite{raissi2018hidden}—a class of methods that integrate neural networks with partial differential equation (PDE) constraints to infer unknown governing equations from data. In these frameworks, the neural network is guided by PDE loss terms (sometimes considered as “soft constraints”), enabling the model to incorporate domain knowledge during training. Such physics-informed approaches often outperform purely data-driven methods when data quality is sufficient, and the targeted PDE is well-characterized, yet their performance can degrade significantly in the presence of sparse or noisy measurements. 
Subsequent extensions to these methods include a variational encoder and hypernetwork to discover stochastic dynamics (HyperSINDy) \cite{jacobs2023hypersindy}, and a Nested SINDy \cite{fiorini2023generalizing} layers that compose symbolic expressions with trainable parameters. Although these approaches can handle modestly complex systems, the studies often focus on low-dimensional or simplified scenarios where the discovered model may be reduced to a single dominant term. Hence, practical challenges remain in scaling them to high-dimensional or highly nonlinear real-world phenomena. 

Another line of research focuses on Bayesian approaches that incorporate uncertainty quantification to address issues related to noisy observations and improve generalizability. For instance, Hirsh et al. \cite{hirsh2022sparsifying} introduced the uncertainty quantification SINDy (UQ-SINDy) framework, which employs sparse Bayesian inference to identify governing equations from data. This method estimates the uncertainty in the values of the model coefficients due to observation errors and limited data, promoting robustness against noise and leading to interpretable models that generalize well. Similarly, the ensemble methods \cite{fasel2022ensemble} handle measurement noise using bootstrap aggregation (bagging), improving the accuracy and robustness of model discovery from extremely noisy and limited data. Specifically, the Ensemble-SINDy (E-SINDy) algorithm identifies an ensemble of SINDy models from subsets of limited data, then aggregates the model statistics to produce inclusion probabilities of the candidate functions, enabling uncertainty quantification and probabilistic forecasts. Additionally, Niven et al. \cite{niven2020bayesian} explored a maximum a posteriori (MAP) Bayesian method for system identification, demonstrating its equivalence to Tikhonov regularization and applying it to the Rössler dynamical system, while Champneys and Rogers \cite{champneys2024bindy} proposed BINDy, a Bayesian approach distinct from previous methods in that it targets the full joint posterior distribution over both the terms in the library and their parameterization in the model, utilizing a Gibbs sampler based on reversible-jump Markov-chain Monte Carlo. However, challenges remain regarding (a) the inclusion or exclusion of critical physics terms and (b) guarantees on the statistical consistency of the inferred terms.


In this work, we theorize \textit{that the observed data from dynamical systems is corrupted not only by measurement noise but also by variability in the system (or input) parameters}. Failing to account for such variability leads to inaccurate inference, resulting in either overfitting or underfitting the governing equations. While the modeling of measurement noise has garnered significant attention,  efforts to account for the effect of input variability on parameter estimation remain largely missing in the context of recovering governing equations. 
Meanwhile, input variability is a common occurrence in real-world systems. An example of input variability can be observed in fluid mechanics, particularly in the process of infiltration. Here, variability in properties such as viscosity and density significantly influences the infiltration rate. Changes in these properties can lead to substantial differences in how fluids penetrate and flow through porous media, thereby affecting the overall efficiency and behavior of the infiltration process \cite{leo2023impact}.
Other examples of input variability include additive manufacturing, where variability in raw materials and process parameters impact the quality of the printed parts \cite{dass2019state, olabiyi2024characterization}, and soil mechanics, where variability in soil moisture and temperature leads to significant alterations in soil infiltration and porosity properties \cite{cheng2024variability}.

To account for the system variability, we re-envision the unknown coefficients as random variables instead of fixed quantities. As such, propagating the variability in these coefficients through the underlying differential equation leads to a push-forward distribution (Equation \eqref{eq:push-forward}) that matches the observed data distribution. The key difference between our formulation and Bayesian inference (e.g., see \cite{hirsh2022sparsifying}) is that in the latter, the posterior distribution is a consequence of the measurement noise in the observed data, which shrinks as more data is gathered, eventually collapsing to a point estimate in the infinite data limit. Hence, traditional methods that only focus on modeling measurement noise do not account for the system variability and therefore fail to accurately recover the physics terms and guarantee consistency of the estimated coefficients. 

To recover the distribution of unknown coefficients within the proposed framework, we adopt the concept of a push-forward measure from probability theory, which quantifies how an input probability measure (distribution over unknown coefficients) is transformed by a mapping (typically governing equations) onto an observation space (observed data distribution). We subscribe to the stochastic inversion approach introduced in \cite{butler2018combining} that enables the estimation of the posterior density of the unknown coefficients using the push-forward measure. We begin by considering sparsity-promoting priors \cite{mitchell1988bayesian, carvalho2009handling} over the unknown coefficients and employ a matched-block bootstrap approach to generate the data distribution when only a single realization from the underlying system is available—a common scenario in real systems. 
The matched-block bootstrap method, as introduced by Carlstein et al. \cite{carlstein1998matched}, enhances the traditional block bootstrap by aligning blocks with matching end values, thereby preserving the dependence structure inherent in time series data such as those generated from dynamical systems. The push-forward measure is computed by simulating the prior samples through the candidate governing equations. A rejection sampling method is proposed to draw samples from the posterior distribution of the coefficients. We subsequently provide the theoretical analysis of the consistency of the posterior push-forward estimates under the bootstrapped data~conditions. 

We validate our methodology using simulated (Lotka--Volterra system with single and multiple sample paths and Lorenz system) and three (3) real-world studies (the Hudson Bay predator-prey system and liquid infiltration systems under different conditions). Our approach successfully identifies the governing equations in all scenarios and accurately infers the parameter distribution despite significant noise (validated using simulated examples). Our method demonstrates superior accuracy in the recovered dynamics and the physics coefficients compared to traditional methods like SINDy \cite{brunton2016discovering} and UQ-SINDy \cite{hirsh2022sparsifying} in accurately learning the governing equations. On average, the proposed method reduces the RMSE of the recovered coefficients by approximately $82\%$ across all case studies, significantly outperforming both SINDy and its Bayesian implementations in recovering governing equations from noisy data.
These results highlight the efficacy of our approach for discovering governing physics in complex systems, applicable in fields like physics, biology,~and~engineering.

The remainder of this paper is structured as follows: The next Section presents the Methodology, which details the proposed physics discovery method, including the solution approaches and theoretical justifications. We apply the method to analyze several examples in the Case Study Section, where we demonstrate the efficacy of our approach. Finally, we conclude with a summary and discuss potential future~directions.

\section{\textbf{Stochastic Discovery of Dynamical Systems}}
\label{Dyn-systm}

The general equation of a dynamical system with state variables $\bm{x}(t) \in \mathbb{R}^p$ is given as, 
\begin{equation}
\dot{\bm{x}}(t) = f(\bm{x}(t), \boldsymbol{\xi})  
\label{dyn-sys}
\end{equation}
\noindent where, $\dot{\bm{x}}(t)$ represents the time derivative of $\bm{x}(t) \in \mathbb{R}^p$ and $p$ is the number of state variables in the system, $  \boldsymbol{\xi} $ are the physics coefficients and $f(\bm{x}(t), \boldsymbol{\xi})$ is the governing physics equation. The physics discovery problem aims to identify the exact form of the function $f(\bm{x}(t), \boldsymbol{\xi})$, i.e., determining the matrix of coefficients $\boldsymbol{\xi}$ of the physics terms from the observed data. A common approach to estimating these coefficients is by expressing the dynamical system as a symbolic regression problem \cite{Brunton2016}, 
\begin{equation}
    \dot{\bm{X}} = \underline{\bm{\lambda}} \Theta(\bm{X}) \label{symb_regress}
\end{equation}
\noindent where $\bm{X} = [\bm{x}(t_1), \ldots, \bm{x}(t_n)]^{\mathsf T}$ is a matrix whose columns corresponds to the time series data observed at $t_1, \dots, t_n$ time steps  from the dynamical system, $\Theta(\bm{X}) \in \mathbb{R}^{m}$ is a vector of $m$ symbolic physics terms (the physics library), some combination of which encodes the dynamics of the system, and 
$\underline{\bm{\lambda}} = [\bm{\lambda}_1, \ldots, \bm{\lambda}_p]^{\mathsf T} \in \mathbb{R}^{p\times m}$ is a sparse matrix containing the unknown coefficients, i.e., the entries of the matrix $\underline{\bm{\lambda}}$ are either $0$ or the coefficients $\lambda_{jk}, \ j = 1, 2, \ldots, p$ and $k = 1, 2, \ldots, m$. The vector $\bm{\lambda}_j \in \mathbb{R}^m$ contains the coefficients of the physics terms corresponding to the $j^{th}$ state variable, with non-zero entries in $\bm{\lambda}_j$ indicating active physics terms in that state variable equation. The library $\Theta(\bm{X})$ is constructed by incorporating prior knowledge of the system's dynamics, such as conservation laws or symmetries, or by systematically including a diverse set of candidate functions to ensure comprehensive coverage of potential governing mechanisms \cite{brunton2016discovering,wang2022sindy}. An example of the library with linear, polynomial, trigonometric, and differential basis terms is,  
\begin{equation}
\Theta(\bm{X}) = 
\begin{bmatrix}
\bm{1} & \bm{X} & \bm{X}^2 & \cdots & \sin(\bm{X}) & \cos(\bm{X}) & \cdots & \dot{\bm{X}} & \ddot{\bm{X}}
\end{bmatrix}
\label{lib-eqn}
\end{equation}
Since the observed data is corrupted with noise, we state the data-generating process as,
\begin{equation}
    \bm{Y} = \int \dot{\bm{X}} \, dt + \bm{\epsilon} = \int  \underline{\bm{\lambda}}\Theta(\bm{X}) dt + \bm{\epsilon},
\label{real-dyn-eq}
\end{equation}
\noindent where $\bm{\epsilon}$ is the measurement noise, and, without loss of generality, is assumed to follow a Gaussian distribution with zero mean and an unknown variance $\sigma^2$, i.e., $\bm{\epsilon} \sim \mathcal{N}(\bm{0}, \sigma^2)$. Here, the measurement noise is assumed to be additive and independent of the actual dynamics.

Initial efforts, e.g., \cite{brunton2016discovering} proposed regularized regression methods such as LASSO to parsimoniously recover the unknown coefficients from measured data \cite{voss2004nonlinear}. More recent works have focused on Bayesian approaches (e.g., \cite{hirsh2022sparsifying}), which model the unknown coefficients as random variables to capture the uncertainty in their estimates due to the measurement noise. Note, however, that this uncertainty reduces as more data becomes available, eventually vanishing in the infinite data limit, in which case the posterior distribution converges to the maximum likelihood estimate. 

Contrary to the existing approaches, we model the unknown coefficients $\underline{\bm{\lambda}} \in \bm{\Lambda}$ as random variables with some probability measure  $\mu_{\bm{\Lambda}}$ that captures their inherent variability. Under this formalism, the physics discovery problem becomes one of inferring the unknown probability measure from observed data. To this end, we borrow the concept of push-forward from probability theory. Within this framework, the push-forward measure---obtained by mapping the true probability measure $ \mu^*_{\bm{\Lambda}}$ through the data-generating process (Equation~\eqref{real-dyn-eq})---matches the empirical probability measure on the observed data. We formalize this by minimizing the Kullback-Leibler (KL) divergence between the push-forward and observed data distribution as, 
\begin{equation}
\label{eq:KL-objective}
\mu^*_{\bm{\Lambda}} = \arg\min_{\mu_{\bm{\Lambda}}} \, D_{KL} \left({\mu}_{\widehat{\bm{Y}}} \,\|\, \mu_{\bm{Y}} \right)
\end{equation}
where $\mu_{\widehat{\bm{Y}}}$ is the push-forward measure and $\mu_{\bm{{Y}}}$ is the empirical measure on the observed data. By minimizing the KL divergence, the push-forward distribution matches the data distribution, making the inferred dynamics consistent with the observed data. In the following, we discuss the approach to inferring the probability measure over the unknown coefficients that capture the variability in input parameters and its regularization based on the prior distribution that shrinks the spurious coefficients to zero, eliminating the irrelevant terms.

\subsection{\textbf{Stochastic Inverse Model}}
\label{ss:Input_var}
For some candidate governing equation $ Q(\underline{\bm{\lambda}}): {\bm{\Lambda}} \to \bm{X} $, the stochastic inverse approach aims to infer a probability measure $\mu_{\bm{\Lambda}}$ on the space of unknown coefficients ${\bm{\Lambda}}$ such that the push-forward of $\mu_{\bm{\Lambda}}$ through $ Q(\underline{\bm{\lambda}})$, combined with additive white noise, matches the probability distribution $\mu_{\bm{Y}}$ of the observed data. To infer $\mu_{\bm{\Lambda}}$, we adopt the push-forward inference proposed in \cite{butler2018combining}. Analogous to standard Bayesian inference, we begin by defining a prior density $\pi_{\bm{\Lambda}}(\underline{\bm{\lambda}})$ over the space of the unknown coefficients $\underline{\bm{\lambda}}$.  Let $\widehat{Q(\underline{\bm{\lambda}})}$ represent a candidate solution for the governing equation $  Q(\underline{\bm{\lambda}})$, and $\pi_{Q}(\widehat{Q(\underline{\bm{\lambda}})})$ be the resulting push-forward density.  Upon observing noisy measurements $\underline{\bm{Y}} = \{\bm{Y}_1, \bm{Y}_2, \ldots, \bm{Y}_r$\}, the posterior density over the unknown coefficients is expressed as, 
\begin{equation}
    \pi_{\bm{\Lambda}}(\underline{\bm{\lambda}}|\underline{\bm{Y}}) 
    \;=\; 
    \pi_{\bm{\Lambda}}(\underline{\bm{\lambda}})
    \;\times\;
    \frac{\pi_{\mathcal{Y}}\left(\widehat{Q(\underline{\bm{\lambda}})}\Big\lvert\underline{\bm{Y}}\right)}{\pi_{Q}(\widehat{Q(\underline{\bm{\lambda}})})},
    \label{eq:push-forward}
\end{equation}

\noindent where $\pi_{\mathcal{Y}}\left(\widehat{Q(\underline{\bm{\lambda}})}\Big\lvert\underline{\bm{Y}}\right)$
is the likelihood that the candidate governing equation $\widehat{Q(\underline{\bm{\lambda}})}$ generated the observed data $\bm{Y}_i \in \mathbb{R}^{p\times n}$. We refer to $\bm{Y}_i$ as a sample path with $n$ being the number of time steps at which data are collected for each of the state variables. 
Note that in many a cases only a single (or very few) sample path(s) is available (i.e., $r=1$). For such instances, we employ a bootstrap procedure (discussed in Section~\ref{subsec:data-distribution}) to generate multiple sample paths from the original data to estimate the likelihood function. We now discuss the multivariate kernel density~estimator~(KDE) to estimate the likelihood function $\pi_{\mathcal{Y}}\left(\widehat{Q(\underline{\bm{\lambda}})}\Big\lvert\underline{\bm{Y}}\right)$ and the push-forward density $\pi_{Q}(\widehat{Q(\underline{\bm{\lambda}})})$.

The KDE approximation is performed for each state variable individually; that is, for every $j = 1,\dots,p$, we collect the sample paths $[\bm{Y}_{i}]_j$  for $i = 1,\dots,r$.  
We fit a Gaussian kernel density estimator with a bandwidth $h$ selected by Scott’s rule of thumb ($h \propto r^{-1/(n+4)}$\,\cite{scott2015multivariate}) or set
manually.  This procedure yields $p$ different KDEs. After approximation, the likelihood of any new trajectory $[\widehat{\bm{Y}}]_{j}$ generated by the candidate governing equation $\widehat{Q(\underline{\bm{\lambda}})}$ 
under the $j^{\text{th}}$ KDE is
estimated as
\[
\widehat \pi_{\mathcal{Y}}\!\left(\widehat{Q(\underline{\bm{\lambda}})}\Big\lvert\underline{\bm{Y}}\right)
  = \frac{1}{r}\sum_{i=1}^{r}
    \exp\!\left(\frac{-\left([\bm{Y}_{i}]_{j}-[\widehat{\bm{Y}}]_j\right)^2}{2h}\right). 
\]

Next, we specify the prior density function $\pi_{\bm{\Lambda}}(\underline{\bm{\lambda}})$ for the unknown coefficients. Since not all terms in the physics library are relevant in defining the governing equation, we employ sparsity-promoting priors that shrink irrelevant coefficients toward zero. We consider two different prior to provide generalizability, (a) spike-and-slab prior~\cite{mitchell1988bayesian,george1993variable} which combines a point mass at zero (the ``spike'') with a diffuse distribution (the ``slab'') to distinguish between irrelevant and relevant coefficients, and (b) regularized horseshoe prior~\cite{carvalho2009handling,piironen2017sparsity} which is a continuous shrinkage prior that aggressively shrinks small coefficients toward zero yet permits larger coefficients to remain unshrunk, thereby managing sparsity in high-dimensional settings.

The spike-and-slab prior is characterized by a combination of a Dirac delta function (``spike'') and a non-degenerate distribution (``slab'') as: 
\begin{equation}
\lambda_{jk} \mid \bm{\beta}_{jk} \sim \bm{\beta}_{jk} \mathcal{N}(0, c^2) + (1 - \bm{\beta}_{jk}) \delta_0,
\end{equation}
where $\lambda_{jk}$ is the coefficient of the $k^\text{th}$ candidate physics for the $j^{\text{th}}$ state variable, $\bm{\beta}_{jk}\sim~\text{Bernoulli}(p)$, and $\delta_0$ denotes the Dirac delta function at zero. This prior follows a hierarchical distribution where it simultaneously (a) allows the coefficients to vary (when $\beta_{jk}= 1$) as per a pre-specified distribution, typically Gaussian with zero mean and variance $c^2$ while (b) shrinking some of the coefficients to zero (when $\beta_{jk}= 0$). 

The regularized horseshoe prior, on the other hand, reduces irrelevant coefficients 
to zero through continuous shrinkage as opposed to hard thresholds using a hierarchical formulation given as,
\begin{align}
\lambda_{jk} \mid \tilde{\beta}_{jk}, \tau \sim \mathcal{N}(0, \tilde{\beta}_{jk}^2 \tau^2), \quad 
\tilde{\beta}_{jk} = \frac{\bm{\beta}_{jk} c}{\sqrt{c^2 + \tau^2 \bm{\beta}_{jk}^2}}
\end{align}
\noindent where $\beta_{jk}\sim\text{C}^+(0,1)$, $\tau \sim \text{C}^+(0, \tau_0)$, and $ c^2 \sim \text{Inv-Gamma}(\nu/2, \nu s^2/2)$.  Here $\text{C}^+(\cdot, \cdot)$ denotes the half-Cauchy distribution, $\text{Inv-Gamma}(\cdot, \cdot)$ denotes the inverse Gamma distribution, and $\nu$ and $ s$ control the slab’s shape. This formulation strikes a balance between sparsity and flexibility: the half-Cauchy priors concentrate mass near zero, pushing irrelevant coefficients to zero while leaving heavy tails for relevant coefficients. Before discussing the posterior sampling procedure, we examine the bootstrapping approach employed in this work to generate multiple sample paths from the observed data.



\subsection{\textbf{Bootstrapping Observed Data}}
\label{subsec:data-distribution}
When actual experiments are affordable to run, it is feasible to observe multiple sample paths $\bm{Y}_1, \bm{Y}_2, \ldots, \bm{Y}_r$, representing distinct samples from the data generating process defined in Equation~\eqref{real-dyn-eq}. These sample paths are used to infer the observed data distribution using the KDE approach discussed in Section \ref{ss:Input_var}, which is subsequently employed to estimate the likelihood function $\pi_{\mathcal{Y}}\left(\widehat{Q(\underline{\bm{\lambda}})}\Big\lvert\underline{\bm{Y}}\right)$. However, in practice, we often cannot collect multiple independent sample paths from real systems due to cost or feasibility constraints. In such situations, we employ a bootstrapping method to derive the likelihood of the candidate governing equation  $\pi_{\mathcal{Y}}\left(\widehat{Q(\underline{\bm{\lambda}})}\Big\lvert\underline{\bm{Y}}\right)$ from the single sample~path.

To this end, we adopt the matched block bootstrap (MBB) technique \cite{carlstein1998matched} that generates multiple sample paths by resampling blocks of contiguous segments with replacement from a single sample path. In this method, the sample trajectory for the $j^\text{th}$ state variable given as $[\bm{Y}]_j = \{y_{j}(t_1), \ldots,\ y_{j}(t_n)\}$ is partitioned into non-overlapping, but contiguous blocks of fixed length $l$, denoted as $B = \{y_j(t'_1), \ldots, y_{j}(t'_l)\}$ where $t'_1, \ldots, t'_l$ is a contiguous time block sampled from $t_1, \ldots, t_n$. After partitioning the entire sample path, we get $n-l+1$ blocks denoted as $B_1, B_2, \ldots, B_{n - l + 1}$. To preserve temporal dependencies, blocks are sampled and matched according to a Markov chain whose transition probabilities are designed to favor blocks with matching endpoints. Specifically, suppose the first $q$ blocks after resampling are $B_{1},\dots,B_{q}$ where $ql<n$, for non-overlapping blocks, the probability that the
\((q+1)^{\text{th}}\) block is some block $B_{j}$ is:

\begin{align}
p(B_q,B_{j}) &\propto
\begin{cases}
\kappa\!\left(\dfrac{B_{q}(t'_l) -B_{j-1}(t'_l)}{h}\right), 
& j \neq 1,\\
\kappa\!\left(\dfrac{B_{(q+1)}(t'_1)-B_{1}(t'_1)}{h}\right), 
& j = 1,\; ql< n,\\
0, 
& j = 1,\; ql = n.
\end{cases}
\label{eq:mbb}
\end{align}

\noindent where $\kappa$ is a symmetric probability density (see \cite{carlstein1998matched}), and $h$ is a bandwidth parameter. Note that the matching procedure in Equation~{\eqref{eq:mbb}} matches the last observation in $B_{q}$ with the last observation in the block preceding $B_{j}$ in the original partitions. 
By iteratively selecting blocks based on these transition probabilities and concatenating them, we construct synthetic sample paths that mimic the statistical properties of the original data (see Theorem 2). We then apply the KDE estimator to the bootstrapped sample paths to estimate $\pi_{\mathcal{Y}}\left(\widehat{Q(\underline{\bm{\lambda}})}|\underline{\bm{Y}}\right)$. We next discuss the posterior sampling procedure. 

\subsection{\textbf{Posterior Distribution Inference via Rejection Sampling}}

To infer the posterior distribution of the unknown coefficients 
$\underline{\bm{\lambda}}$, we adopt a rejection sampling approach. 
Algorithm~\ref{algorithm} summarizes how this ratio is used to construct the posterior distribution of $\underline{\bm{\lambda}}$. In summary, $ N$ independent samples $ \{\underline{\bm{\lambda}}_i\}_{i=1}^N$ of the unknown coefficients are drawn from the prior distribution $ \pi_{\bm{\Lambda}}(\underline{\bm{\lambda}})$. For each sample $ \underline{\bm{\lambda}}_i$, we simulate the dynamics via the candidate governing equation $ \widehat{Q(\underline{\bm{\lambda}}_i)}$ and collect the resulting sample paths. These simulated paths are then used to approximate the distribution $ \pi_{Q}\left(\widehat{Q(\underline{\bm{\lambda}})}\right)$ using the KDE method described above. 

The rejection sampler for the sampled coefficients $\underline{\bm{\lambda}}$ is defined 
using the ratio of the likelihood function 
$\pi_{\mathcal{Y}}\left(\widehat{Q(\underline{\bm{\lambda}})}|\underline{\bm{Y}}\right)$ and the push-forward of the prior distribution $\pi_{Q}(\widehat{Q(\underline{\bm{\lambda}})})$  as,

\begin{equation}
    \varphi\left(\widehat{Q(\underline{\bm{\lambda}})}\right) 
    \;=\;
    \frac{
      \pi_{\mathcal{Y}}\left(\widehat{Q(\underline{\bm{\lambda}})}|\underline{\bm{Y}}\right)
    }{
      \pi_{Q}(\widehat{Q(\underline{\bm{\lambda}})})
    }
    \label{eq:transform}
\end{equation}
which quantifies how well sample paths from the candidate governing equation align with the observed data. 
Any sample whose $\varphi$-value, after normalization by the global maximum $M$ (defined in Algorithm \ref{algorithm}), exceeds a uniform random draw is \emph{accepted} into the posterior sample set. Through repeated evaluations, the method preferentially retains samples that match the data distribution while rejecting those that are inconsistent with observed system behavior. 

\begin{algorithm}[htbp]
\caption{Rejection Sampling}
\normalsize
\begin{algorithmic}[1]
\State \textbf{Input:} Prior samples $\left\{\underline{\bm{\lambda}}_i\right\}_{i=1}^{N}$, total~sample size~$N$
\State \textbf{Compute} $\left\{\widehat{Q({\underline{\bm{\lambda}_i}})}\right\}_{i=1}^{N}$ and $\left\{\varphi\left(\widehat{Q({\underline{\bm{\lambda}_i}})}\right)\right\}_{i=1}^{N}$
\State \textbf{Compute} $M \gets \max_{i}\;\varphi\left(\widehat{Q({\underline{\bm{\lambda}_i}})}\right)$
\For{$i \gets 1 \textrm{ to } N$}
    \State Extract $\varphi\left(\widehat{Q(\underline{\bm{\lambda}}_i)}\right)$
    \State Generate $\xi_i \sim \mathrm{Uniform}(0, 1)$
    \State Compute $\eta_i \gets \varphi\left(\widehat{Q(\underline{\bm{\lambda}}_i)}\right) \,/\,M$
    \If{$\eta_i \;\ge\; \xi_i$}
        \State \textbf{accept} $\underline{\bm{\lambda}}_i$
    \Else
        \State \textbf{reject} $\underline{\bm{\lambda}}_i$
    \EndIf
\EndFor
\State \textbf{Output:} Accepted samples approximating the posterior $\pi_{\bm{\Lambda}}(\underline{\bm{\lambda}}|\underline{\bm{Y}})$
\end{algorithmic}
\label{algorithm}
\end{algorithm}

\noindent
The normalization constant $ M$ in Algortihm \ref{algorithm} ensures that all ratios 
$\eta_i \le 1$. 
Moreover, by linking the data-driven likelihood 
$\pi_{\mathcal{Y}}\left(\widehat{Q(\underline{\bm{\lambda}})}|\underline{\bm{Y}}\right)$ to the prior push-forward density $\pi_{Q}\left(\widehat{Q(\underline{\bm{\lambda}})}\right)$, 
the algorithm enforces consistency with both the prior belief about $\underline{\bm{\lambda}}$ and the observed data. As more data become available, $\pi_{\mathcal{Y}}\left(\widehat{Q(\underline{\bm{\lambda}})}|\underline{\bm{Y}}\right)$ becomes increasingly localized, thereby refining the set of accepted samples toward a posterior distribution 
$\pi_{\bm{\Lambda}}(\underline{\bm{\lambda}}|\underline{\bm{Y}})$ 
that most accurately reflects the unknown dynamical system. 

In high-dimensional settings, the acceptance rate may be very low if 
$\varphi(\cdot)$ spans a large range of values, or if the data are extremely informative. In such cases, one can employ additional sampling strategies (e.g., importance sampling or Markov chain Monte Carlo) to further improve the efficiency of posterior inference. 

\subsection{\textbf{Statistical Consistency under Bootstrapping}}

The matched block bootstrap approach discussed in Section~\ref{subsec:data-distribution} offers a statistically consistent way to estimate the probability measure $\widehat{\mu}_{\bm{Y}}$ by resampling multiple sample paths that preserves the temporal dependencies. This is the key to accurately estimating  the likelihood function $\pi_{\mathcal{Y}}\left(\widehat{Q(\underline{\bm{\lambda}})}\Big\lvert\underline{\bm{Y}}\right)$. The consistency of MBB estimator is elaborated in Theorem 1.





\begin{theorem}[Consistency of the MBB]
Let $\bm{Y}'_{1},\bm{Y}'_{2},\ldots,\bm{Y}'_{r}$ be $r$ block-resampled replicates of the observed time series $\bm{Y}$, each with finite variance.  
Denote by $\mu_{\bm{Y}}$ the true distribution of $\bm{Y}$ and by $\mu_{\bm{Y}^{*}}$ the distribution of a \emph{generic} bootstrap replicate
$\bm{Y}^{*}\in\{\bm{Y}'_{1},\ldots,\bm{Y}'_{r}\}$.  
Then the empirical bootstrap distribution $\widehat{\mu}_{\bm{Y}}$ is consistent in the sense that
\[
    d_{K}\bigl(\mu_{\bm{Y}},\mu_{\bm{Y}^{*}}\bigr)
    \;\xrightarrow[\;r\to\infty\;]{\text{a.s.}}\;0,
\]
where
   $ d_{K}(\mu,\nu)
    \;=\;
    \sup_{\bm{Y}\in\mathbb{R}^{p\times n}}
    \bigl|
        \mu\bigl((-\infty,\bm{Y}]\bigr)
        -
        \nu\bigl((-\infty,\bm{Y}]\bigr)
    \bigr|$
is the \emph{Kolmogorov metric} (uniform distance) between distribution functions.
\end{theorem}

In other words as the number of sample paths generated by the MBB approach increase, the distance between the true and bootstrapped probability measure over the observed data reduces to zero (refer to \cite{carlstein1998matched} for the proof). 
We next show that the posterior distribution of the unknown coefficients inferred via the stochastic inverse approach is a consistent solution of the physics discovery problem. In other words, if the samples of  $\underline{\bm{\lambda}}$ drawn from the posterior distribution are propagated through the recovered governing equations, the resulting push forward distribution would match that of the observed data.

\begin{theorem}[Consistency of push-forward]

\label{thm:push-forward_consistency} 
The posterior probability measure over the unknown coefficients given as, 
\begin{equation}
  \pi_{\bm{\Lambda}}(\underline{\bm{\lambda}}|\underline{\bm{Y}}) = 
\int_{\underline{\bm{Y}}} \left(
  \int_{A \cap Q^{-1}(q)} 
    \pi_{\bm{\bm{\Lambda}}}(\underline{\bm{\lambda}}) \, 
    \frac{\pi_{\mathcal{Y}}\left(\widehat{Q(\underline{\bm{\lambda}})}|\underline{\bm{Y}}\right))}{\pi_{Q}(\widehat{Q(\underline{\bm{\lambda}})})} 
    \,d\tilde{\mu}_{{\bm{\Lambda}}, q}(\underline{\bm{\lambda}})
\right) \,d\tilde{\mu}_{\mathcal{Y}}(q)
\end{equation}where $q = Q(\underline{\bm{\lambda}})$, $\tilde{\mu}_{\mathcal{Y}}$ and $\tilde{\mu}_{\Lambda, q}$ are volume measures  is a consistent solution to the stochastic inverse problem in the sense that for some measurable set $A$ in the metric space $\mathcal{Y}$, the push-forward $\pi_{Q}(\widehat{Q(\underline{\bm{\lambda}})})$ for any $\underline{\bm{\lambda}}$ $\in \pi_{\bm{\Lambda}}(\underline{\bm{\lambda}}|\underline{\bm{Y}})$ satisfies,  

\begin{equation}
\mu_{\Lambda}(\hat{Q}^{-1}(A)) = \mu_{\mathcal{Y}}^{\widehat{Q(\mu_{\Lambda})}}(A) = \mu_{\mathcal{Y}}(A)
\end{equation}

\noindent where $\widehat{Q^{-1}(A)} = \{\underline{\bm{\lambda}} \in \Lambda \mid \widehat{Q(\underline{\bm{\lambda}})} \in A)\}$ meaning that the probability distribution of the observed data $\pi_{\mathcal{Y}}(A)$ matches the distribution of the data generated by the posterior samples, i.e., $\pi_{\bm{\Lambda}}(\underline{\bm{\lambda}}|\underline{\bm{Y}})$ when propagated through the recovered model $\widehat{Q(\underline{\bm{\lambda}})}$.






\end{theorem}

Refer to \cite{butler2020data} for details and proof. These two results combined ensure the consistency of the push-forward of the posterior computed from the bootstrapped data.

\section{\textbf{Results}}
\label{sec:case_study}

In this section, we demonstrate the performance of our method by applying it to three different dynamical systems: (a) Lotka--Volterra, (b) Lorenz, and (c) infiltration system. We utilized both simulated and real experimental datasets for these systems and briefly discussed them in the following.
We assess the performance of the method in comparison with two other state-of-the-art methods: SINDy \cite{brunton2016discovering} and  Bayesian (UQ-SINDy) approach presented in \cite{hirsh2022sparsifying}. The summary of results for all case studies considered in this work is presented in Table \ref{summaries}. 

\begin{table*}[ht]
    \centering
    \caption{Comparison of recovered governing equations across all case studies, with root mean square errors (RMSE) calculated relative to the ground truth model coefficients in each scenario}
    \includegraphics[width=1.0\textwidth]{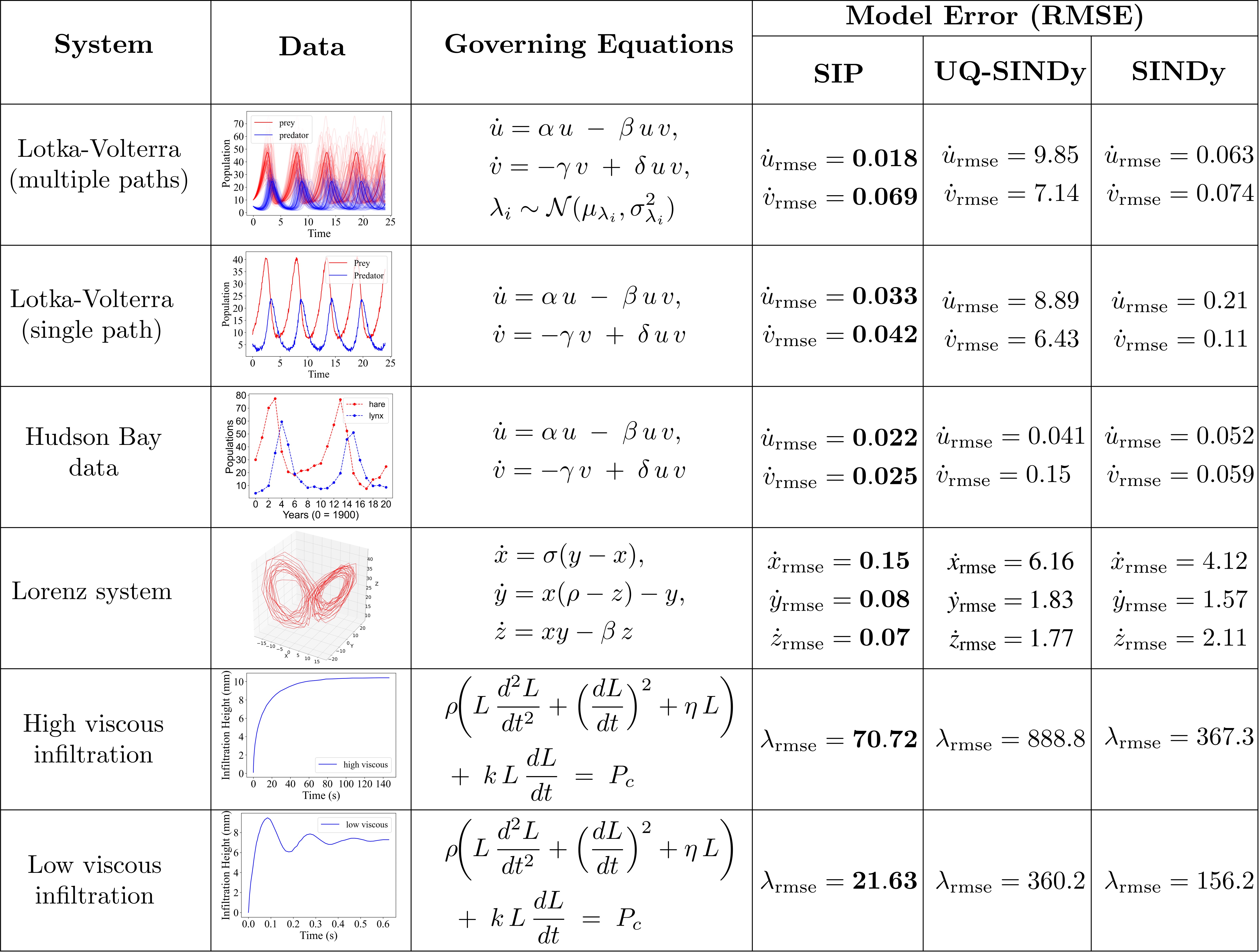}
    \label{summaries}
\end{table*}

\subsection{\textbf{Lotka--Volterra System}}
\label{cs:lv_model}

The Lotka--Volterra system, also known as the predator-prey model, describes the dynamics of ecological systems in which two species interact, one as a predator and the other as prey. The governing equations representing this coupled system are defined as,
\begin{equation}
\begin{aligned}
\dot{u} &= \alpha u - \beta uv, \\
\dot{v} &= - \gamma v + \delta uv 
\end{aligned}
\label{lv-eqn}
\end{equation}
where $  u $ and $  v $ represent the prey and predator populations, respectively, and $  \alpha, \beta,\delta, \gamma $ are the unknown coefficients. The objective is to recover the underlying dynamics of the Lotka--Volterra system across three distinct scenarios: a simulated case with input variability (i.e., coefficients are random variables), a simulated single-trajectory with input variability and additive noise (demonstrating the application of the matched-block bootstrap technique), and the Hudson Bay dataset \cite{hewitt1921conservation} representing a natural Lotka--Volterra system. We discuss each of the cases in the following. 

\subsection*{\textbf{Simulated Lotka--Volterra}}

For the Lotka--Volterra system with input variability, we define each coefficient as a random variable drawn from a normal distribution as,
\[
\alpha \sim \mathcal{N}(1.0, 0.1^2), \quad
\beta \sim \mathcal{N}(0.1, 0.01^2), \quad
\gamma \sim \mathcal{N}(1.5, 0.15^2), \quad
\delta \sim \mathcal{N}(0.075, 0.0075^2).
\]
We sampled 100 coefficient combinations from these distributions and simulated the model. To introduce additive noise, we first compute the \emph{root‑mean‑square} (RMS) amplitude of the simulated noiseless data $\bm{Y}$,
\[
\mathrm{RMS}(\bm{Y}) \;=\; \frac{1}{N} \sum_{i=1}^{N} \bm{Y}_i^{2},
\]
and set the noise standard deviation to $\sigma \;=\; 0.1\,\mathrm{RMS}(\bm{Y})$. Here, $\sigma$ denotes the standard deviation of the zero‑mean Gaussian noise added to each measurement, and this choice yields a signal‑to‑noise ratio (SNR) of approximately $ 20\,\mathrm{dB}$.

For the single-path data, we simulate the Lotka--Volterra system using the mean parameter values. We then superimpose a lognormal \emph{multiplicative} noise with $\mu = 0$ and $\sigma = 0.1$ on the simulated data. Here, $\sigma = 0.1$ indicates roughly a $ 10\%$ noise level relative to the signal amplitude, also giving an SNR of about $ 20\,\mathrm{dB}$. 
The simulated time series (covering about four oscillation periods) are shown in Table~\ref{summaries}, rows 1 and 2. We apply the three methods to these datasets to recover the governing physics and the distribution of the coefficients for the corresponding physics~terms.

\begin{figure}[!t]
    \centering
    \includegraphics[width=0.8\textwidth]{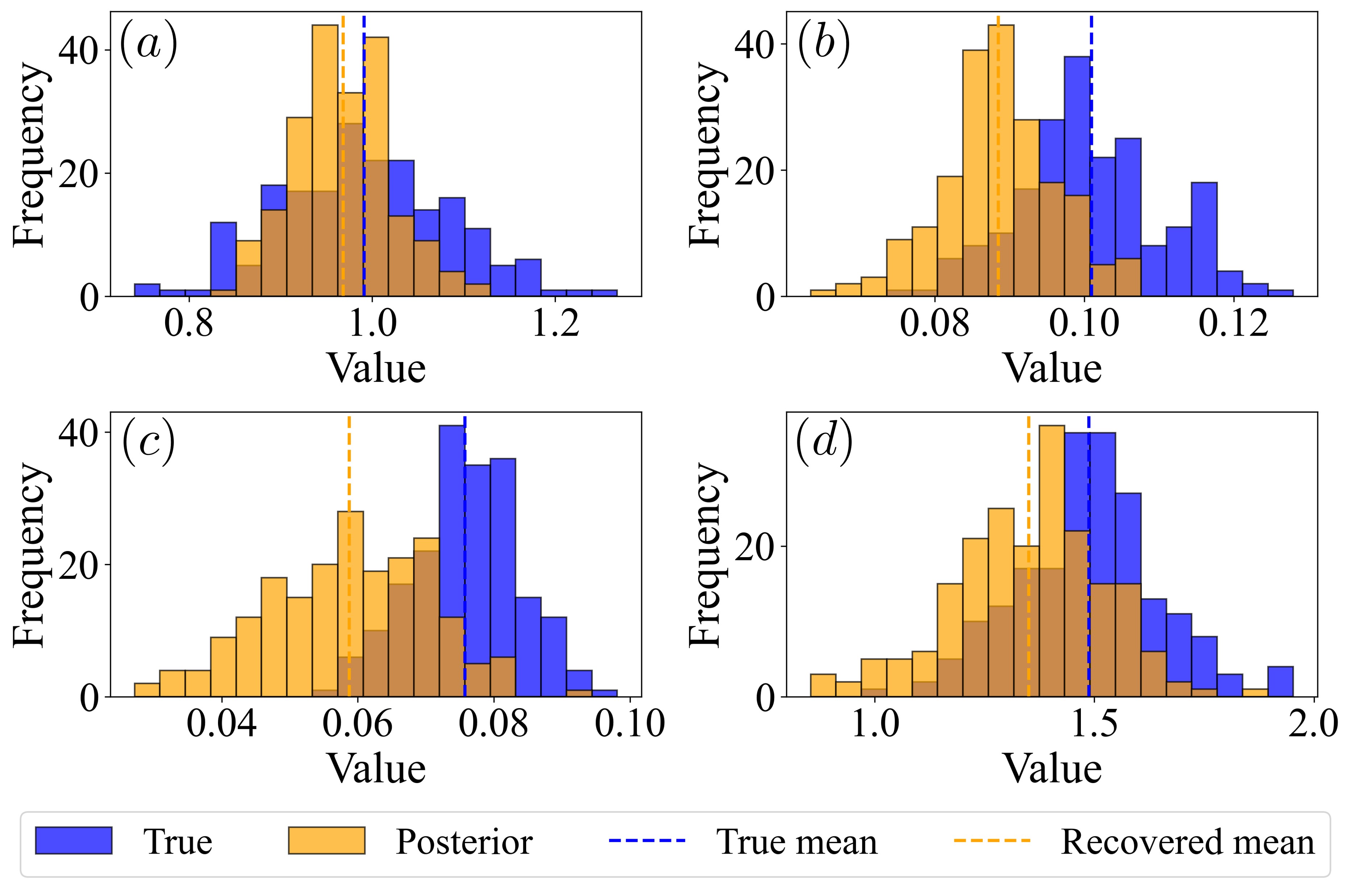}
    \caption{Comparing the true and recovered distributions of  (a) $\alpha$ (b) $\beta$ (c) $\delta$ and (d) $\gamma$ parameters for the Lotka--Volterra system with input variability.}
    \label{Lotka_voltera_sip_param_dist}
\end{figure}

Compared to the stochastic inversion approach, both SINDy and UQ-SINDy were less accurate in capturing the true parameter distributions. SINDy recovered a deterministic parameter set that includes additional spurious terms and more significant overall errors, as evidenced in Table~\ref{summaries} (rows 1 and 2). The UQ-SINDy approach did produce posterior distributions, but its recovered means deviated more significantly from the true values, and the variances were generally higher, indicating less confident estimates. Overall, the proposed method not only identifies the correct terms but also recovers posterior distributions that closely match the ground truth (Figure~\ref{Lotka_voltera_sip_param_dist}), as indicated by an average KL divergence of about 0.075. In the multi-path scenario, SIP achieves roughly 99\% lower RMSE than SINDy and up to 70\% lower than UQ-SINDy, while in the single-path scenario, improvements remain similarly high. These findings underscore the method’s robustness, balancing accurate model recovery with reliable coefficient distribution estimation even under noisy conditions.

\subsection{\textbf{Lynx-hare model}}
\label{sec:hudson_bay_dataset}
\begin{figure}[htbp]
    \centering
    \includegraphics[width=0.9\columnwidth]{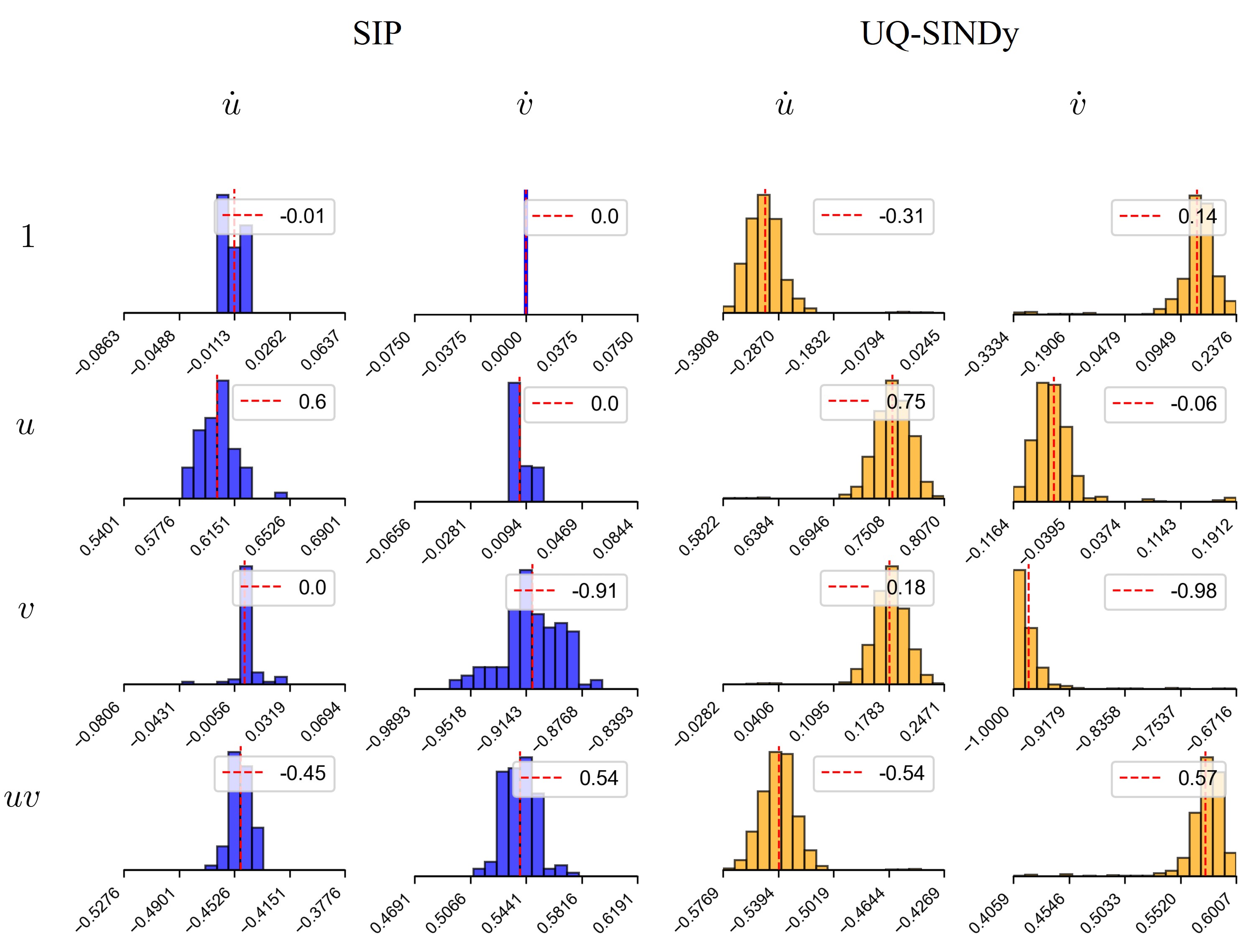}
    \caption{Posterior marginal distributions of the Lotka--Volterra coefficients inferred from the Hudson Bay data.  
             Blue histograms (left block) are from the SIP solution, and orange histograms (right block) are from UQ--SINDy.  Within each method, the two columns correspond to the coefficients of $\dot u$ (left) and $\dot v$ (right); rows list the candidate library terms $1$, $u$, $v$, and $uv$.  }
    \label{HB_param_dist}
\end{figure}

The Hudson Bay dataset is a historically significant ecological dataset capturing long-term lynx–hare predator-prey cycles. Its clear cyclical patterns make it an exemplary reference for understanding population dynamics and ecological interactions under real-world conditions, thereby offering invaluable scientific insight into how species populations fluctuate over extended timescale \cite{hewitt1921conservation}. The dataset comprises the yearly records of lynx and hare pelts collected by the Hudson Bay Company over a span of 20 years. The pelt counts serve as a proxy for the populations of lynx and hares, offering valuable insights into the predator-prey dynamics characteristic of these species.
 By leveraging the proposed method, we recovered the Lotka--Volterra dynamics governing this data (see Table \ref{summaries} row 3).  The result presented in Figure \ref{HB_param_dist} shows the recovered distribution of coefficients of the library terms. 
 Compared to the proposed method, both SINDy and UQ-SINDy struggled to accurately identify the core physics terms, as indicated by the additional spurious terms they inferred, as shown in Table \ref{summaries} row 3. SINDy does not produce a coefficient distribution for uncertainty analysis. Although the UQ-SINDy approach does provide distributions of the unknown coefficients, it exhibits higher variance and systematic bias away from the ground truth (Figure~\ref{HB_param_dist}), suggesting that the recovered model may be less accurate under noisy conditions. On average, our method yields lower RMSE values (Table~\ref{summaries}, row~3), indicating more precise coefficient estimates and stronger agreement with the observed predator--prey dynamics. In particular, SIP reduces the RMSE by over $45\%$--$80\%$ relative to SINDy and by about $58\%$ compared to UQ-SINDy, thereby providing a more faithful reconstruction of the lynx--hare interactions.

\begin{figure}[htbp]
    \centering
    \includegraphics[width=\columnwidth]{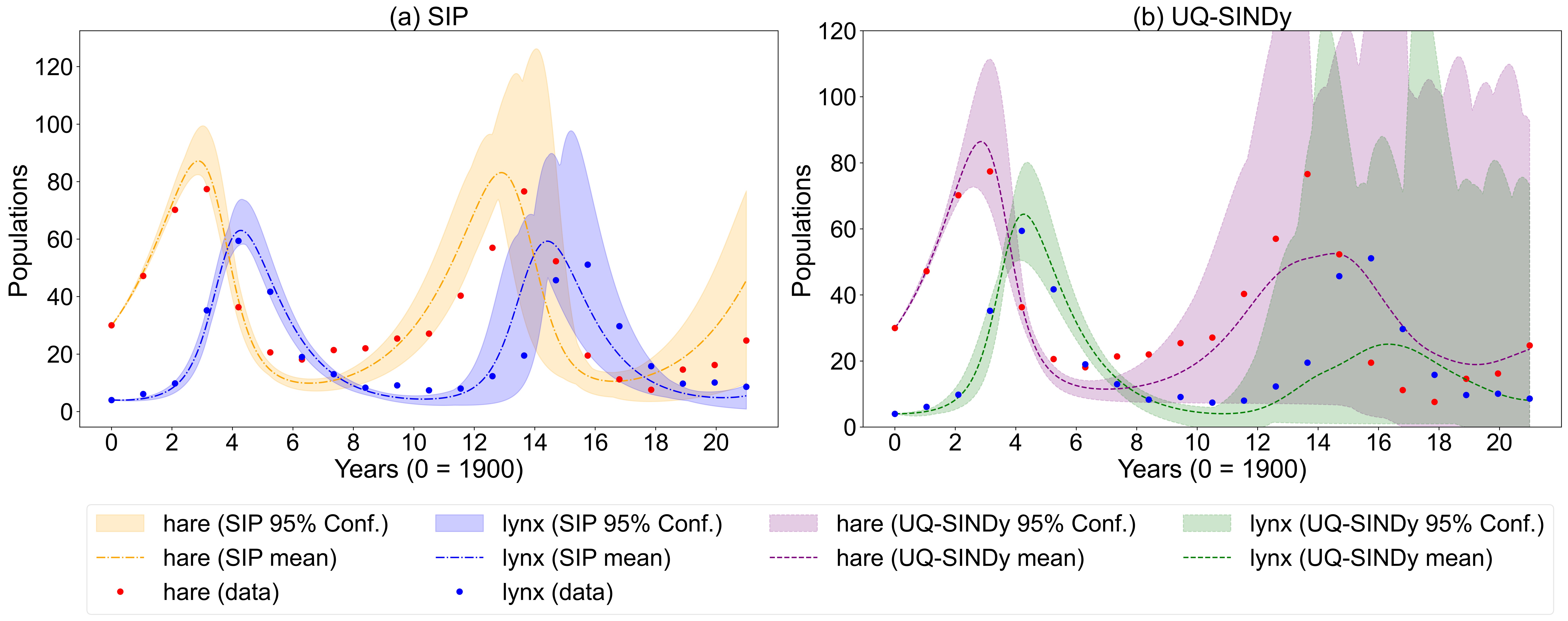}
    \caption{Posterior predictive hare–lynx trajectories for the Hudson Bay data using (a)~SIP and (b)~UQ--SINDy.  
    Dashed lines show posterior means, shaded regions the 95\% credible intervals, and dots the observed populations
    (year 0 = 1900).}
    \label{HB_lv_post}
\end{figure}

We also performed a posterior predictive analysis by computing the push-forward of the recovered distribution to generate a confidence interval on the observed data, as shown in Figure~\ref{HB_lv_post}. These posterior predictive plots reveal how effectively the recovered model captures the amplitude and timing of the lynx–hare population cycles while also illustrating the extent of our uncertainty. Notably, most observed data points lie within the 95\% credible bounds (the shaded regions), indicating that the inferred model coefficients yield predictions consistent with historical records. Moreover, slight deviations in peak magnitudes or trough depths highlight the regions where the model or parameter uncertainties may be higher, suggesting potential avenues for model refinement or additional data collection. Overall, this predictive interval analysis underscores both the strengths and limitations of the inferred dynamics, demonstrating reasonable agreement with real-world observations and providing a quantitative window into how confidently we can forecast the populations given the recovered physical model and data.

\subsection{Lorenz System}
The Lorenz system is a canonical model for chaotic dynamics, originally developed to represent atmospheric convection. The system is defined by a set of three coupled nonlinear differential equations, as shown in the fourth row of Table \ref{summaries},
where $\sigma$, $\rho$, and $\beta$ are parameters representing the Prandtl number, Rayleigh number, and a geometric factor, respectively. The Lorenz system exhibits rich and chaotic behavior, characterized by sensitive dependence on initial conditions, which is a hallmark of chaotic systems. This sensitivity leads to divergent sample paths over time despite starting from nearly identical initial conditions. This system is an excellent benchmark for testing the efficacy of physics identification methods due to its inherent complexity and chaotic behavior \cite{brunton2016discovering}.

For this case study, we simulate the Lorenz system with $\sigma=10$, $\rho=28$, and $\beta=\frac{8}{3}$. A 10\% additive noise is introduced, corresponding to an SNR of roughly 20 dB, thereby producing noisy sample paths for the subsequent physics discovery problem. As summarized in Table~\ref{summaries} (row 4),  the SINDy method misidentifies specific terms under these noisy conditions, while UQ-SINDy provides broader posterior estimates that deviate more from the actual coefficients. In contrast, the stochastic method closely reconstructs the Lorenz equations despite the chaotic nature and measurement noise, accurately recovering each governing term’s coefficients and showing less divergence between recovered and true parameters. In terms of RMSE, SIP achieves improvements of roughly $95\%$--$98\%$ over SINDy and $90\%$--$96\%$ over UQ-SINDy (Table~\ref{summaries}, row~4), underscoring its superior ability to recover the chaotic Lorenz dynamics more accurately. This robustness in the presence of both chaos and noise further highlights the advantage of the proposed method regarding structural consistency and parameter accuracy.

\begingroup
  \setlength{\jot}{8pt}%
  \renewcommand{\arraystretch}{1.4}%

  \begin{table}[htbp]
  \centering
  \caption{Recovered dynamics of different methods for the Lorenz system.  
  The mean of the recovered distributions was used to represent each coefficient in both SIP and UQ‑SINDy.}
  \label{learned_lorentz}
  \resizebox{\textwidth}{!}{%
    \begin{tabular}{|c|c|c|c|}
      \hline
      \textbf{True Lorenz dynamics} 
        & \multicolumn{3}{c|}{\textbf{Recovered Lorenz dynamics}}\\
      \cline{2-4}
        & \textbf{SIP recovered} 
        & \textbf{SINDy} 
        & \textbf{UQ SINDy}\\
      \hline
      \(
        \begin{aligned}
          \dot{x}&=-10x+10y\\
          \dot{y}&=28x - xz - y\\
          \dot{z}&=xy - 2.67z
        \end{aligned}
      \)
      &
      \(
        \begin{aligned}
          \dot{x}&=-9.85x+9.86y\\
          \dot{y}&=29.3x - 0.91xz - 0.87y\\
          \dot{z}&=1.07xy - 2.74z
        \end{aligned}
      \)
      &
      \(
        \begin{aligned}
          \dot{x}&=-9.47x+9.47y\\
          \dot{y}&=0.322 + 22.3x - 0.84xz\\
          \dot{z}&=-1.75 + 0.94xy - 2.46z
        \end{aligned}
      \)
      &
      \(
        \begin{aligned}
          \dot{x}&=-9.77x+9.59y\\
          \dot{y}&=0.58 + 24.7x - 0.7xz - 0.81y\\
          \dot{z}&=-0.21 + 1.14xy - 2.38z
        \end{aligned}
      \)
      \\
      \hline
    \end{tabular}%
  }
  \end{table}

\endgroup

\subsection{Infiltration physics}

In this section, we investigate the infiltration process in porous media using experimental data with different viscosity levels \cite{pandey2024multimodal}. The objective is to uncover the underlying physics governing the infiltration process under varying conditions and recover the posterior distribution of the coefficients of the identified terms in the library. We employ two different infiltration datasets: (a) ether, a low-viscosity fluid (0.89\,mPa\,s) whose infiltration is largely driven by inertial forces, and (b) silicone oil, a high-viscosity fluid (0.5\,Pa\,s) where viscous damping plays a dominant role. Ether, with its notably low viscosity, offers a distinct perspective on the dynamics of fluid rise in capillary tubes, predominantly governed by inertial effects rather than viscous damping (see Table \ref{summaries} row six). Conversely, silicone oil, known for its high viscosity, provides a unique opportunity to explore the dynamics of fluid rise in capillary tubes, significantly influenced by viscous forces. This study used the observed data from the infiltration experiments employing silicone oil and Ether, presented in \cite{quere1997inertial}, to recover the infiltration governing equations. 

\begin{figure}[htbp]
    \centering
    \includegraphics[width=0.85\textwidth]{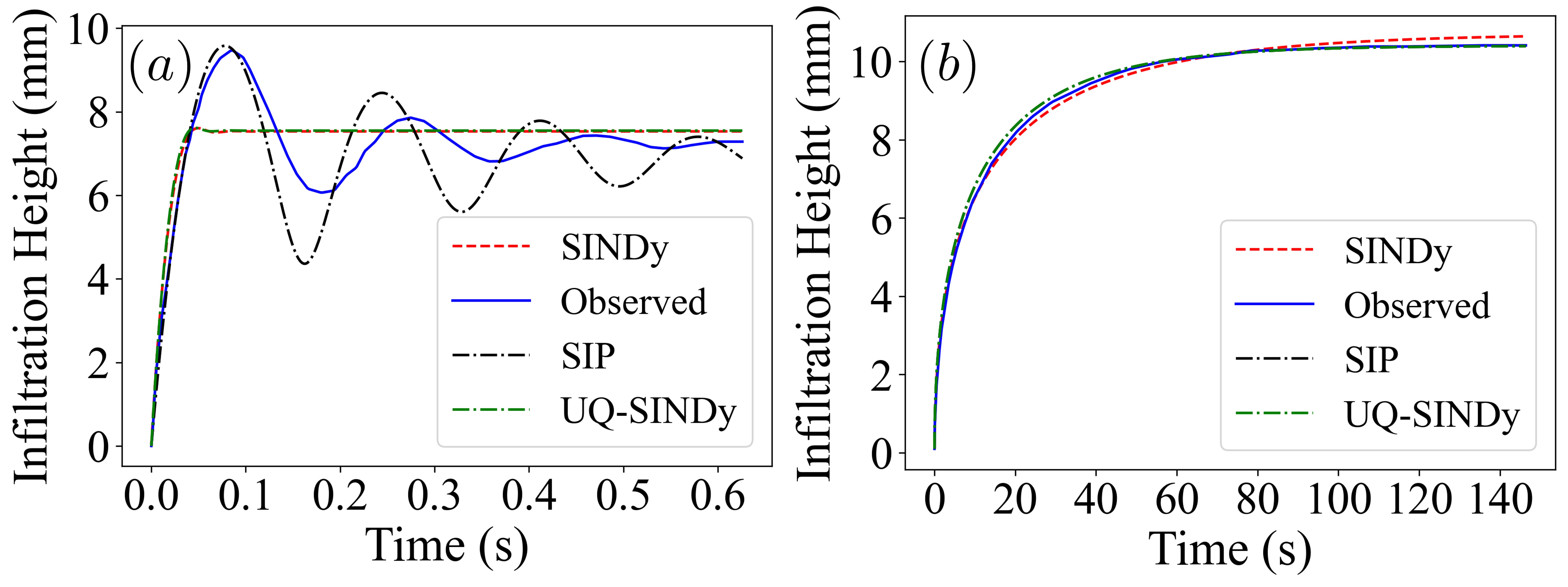}
    \caption{Comparisons of predicted dynamics for (a) low viscous (b) high viscous infiltration system.}
    \label{Low_viscous_fluid_height_comp_base}
\end{figure}

Given that the infiltration physics has been thoroughly studied in the literature, we know the actual governing physics of the infiltration system and the values of the parameters of the fluid is also known from \cite{quere1997inertial} (see Table~\ref{summaries}, rows~5--6). This enables a direct comparison of recovered parameters to true values.  From the RMSE values in the Table, it is clear that the proposed method outperforms both SINDy and UQ-SINDy in capturing the infiltration dynamics. Specifically, for the high‐viscosity fluid, SIP achieves improvements of about $92\%$ and $80.8\%$ over SINDy and UQ-SINDy, respectively; for the low‐viscosity fluid, RMSE drops from $360.19$ and $156.19$ to $21.63$, yielding improvements of approximately $94\%$ and $86\%$. While SINDy fails to capture important physical terms in both regimes and UQ-SINDy includes extraneous terms and overestimates certain coefficients, SIP more accurately models the interplay of viscous and inertial influences. Indeed, as illustrated in Figures~\ref{Low_viscous_fluid_height_comp_base}(a) and~\ref{Low_viscous_fluid_height_comp_base}(b), the simulated sample paths using the recovered models from SIP match the observed data more closely than those of the competing methods—particularly for the low‐viscosity ether, where inertial effects dominate. This stronger agreement underscores the suitability of the proposed learning framework for identifying correct physical terms and accurately estimating their coefficients in both high‐ and low‐viscosity infiltration scenarios.

\section{Conclusion}
\label{sec:conclusion}

We developed a novel stochastic inversion approach for discovering the governing physics equations of complex dynamical systems from observed data. We argued that the observed data from dynamical systems is corrupted with
measurement noise and by the variability in the system (or input) parameters. Failing
to account for such variability--commonly noted in state-of-the-art methods---leads to inaccurate inference, resulting in either overfitting or underfitting the governing equations. To recover the governing equations under this formalism, we introduced a push-forward-based inference, that allows us to account for both inherent system variability and measurement noise in the observed data during the inference step. We provide a Monte Carlo sampling approach to sample from the posterior distribution and discuss the consistency of the proposed stochastic inverse method. Validated on a range of simulated and real-world case studies including Lotka--Volterra predator--prey interactions, the chaotic Lorenz system, and high/low-viscosity infiltration processes, our approach consistently outperforms established techniques mainly SINDy and UQ-SINDy. Notably, in each scenario, SIP yields considerably lower root mean square error (RMSE) values (often exceeding 90\% improvements) with respect to the recovered coefficients, providing a more accurate reconstruction of both the model structure and the coefficient distributions.  
Future extensions include disentangling system variability from measurement noise more explicitly and developing advanced sampling procedures to handle high-dimensional problem settings, further enhancing the power and versatility of the proposed framework.

\FloatBarrier

\bibliographystyle{ieeetr}
\bibliography{main} 

\end{document}